\documentclass[letterpaper, 10 pt, conference]{ieeeconf} 

\IEEEoverridecommandlockouts                              

\usepackage{censor}
\usepackage{graphicx}
\usepackage{booktabs}
\usepackage{amsmath}
\usepackage{tabularx}
\usepackage{array}  
\usepackage{amssymb}
\usepackage{cuted}
\usepackage{listings}
\usepackage{xcolor}
\usepackage{dblfloatfix}
\usepackage{listings}
\usepackage{caption}
\usepackage[noadjust]{cite}
\usepackage{caption}
\usepackage{url}
\title{\LARGE \bf
AERMANI-PLACE: Language Guided Object Placement with Aerial Manipulators
}

\author{%  
    {Sarthak~Mishra$^{*1}$, Ritama Sanyal$^{*1}$, Rishabh~Dev~Yadav$^{2}$, Wei~Pan$^{3}$,~and~Spandan~Roy$^{1}$}%
    \thanks{$^{*}$Equal contribution.}%
    \thanks{$^{1}$Robotics Research Center, IIIT Hyderabad, India.
    Emails: \texttt{\{sarthak.mishra,ritama.sanyal\}@research.iiit.ac.in}, \texttt{spandan.roy@iiit.ac.in}}%
    \thanks{$^{2}$Department of Computer Science, University of Manchester, UK.
    Emails: \texttt{rishabh.yadav@postgrad.manchester.ac.uk}}
    \thanks{$^{3}$ Newcastle University, UK
    Emails: \texttt{wei.pan2@newcastle.ac.uk}}%
}

\begin{document}
\maketitle

\begin{strip}
    \centering
    \vspace{-27mm} % Adjust this to close the gap with the title
    \includegraphics[width=\textwidth]{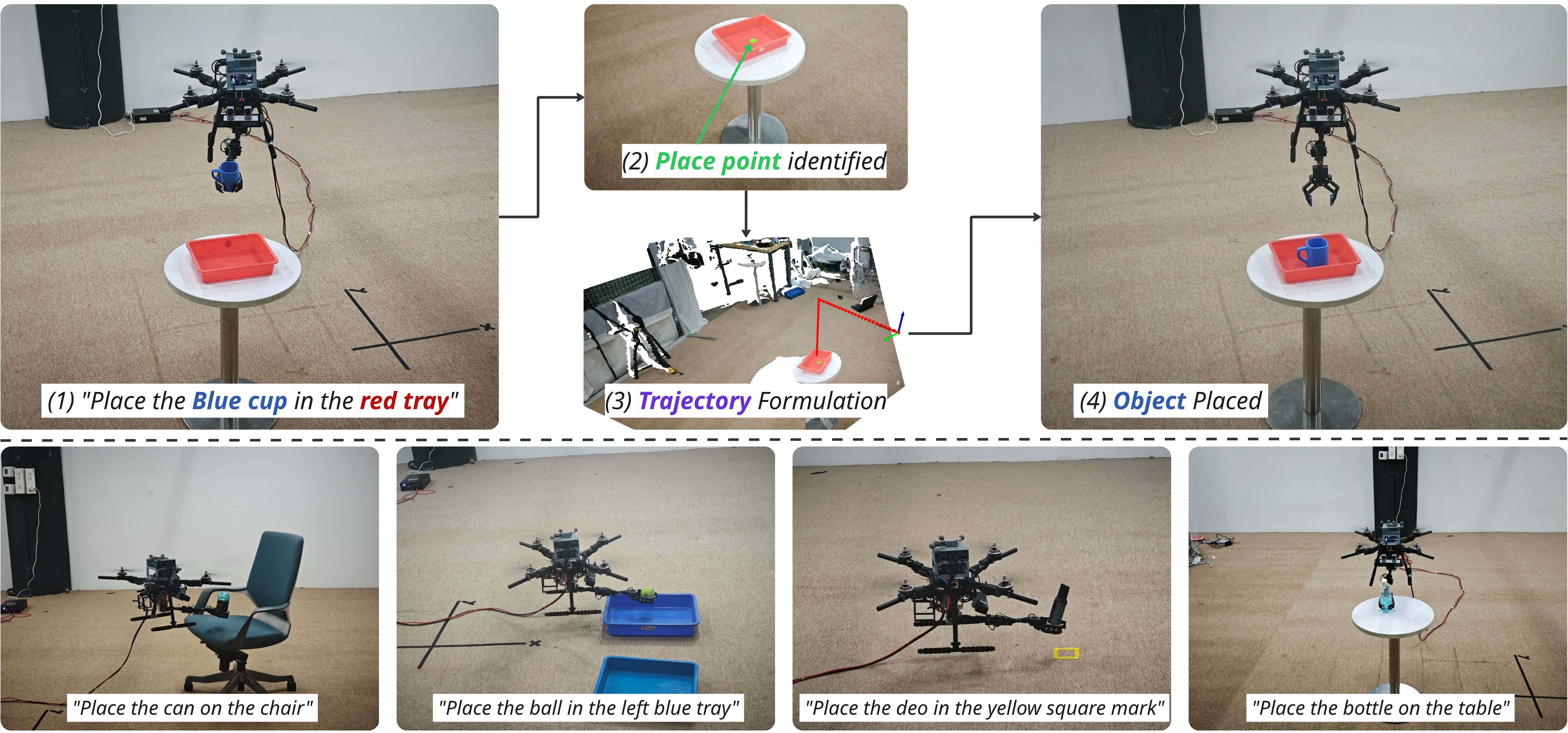}
%     \captionof{figure}{\small \textbf{Overview of AERMANI-PLACE.}
% Given a natural language instruction and RGB-D observations of the placement scene, our method infers the placement location and executes the task using an Aerial Manipulator(AM). 
% (1) {Instruction and Scene Observation:} The AM observes the environment and receives a language instruction specifying the desired placement task. 
% (2) {Placement Marker Prediction:} An image editing model predicts a visual marker in the scene image indicating where the object should be placed. 
% (3) {3D Placement Recovery and Trajectory Formulation:} The predicted marker is projected into the scene using depth observations to recover a metric placement point, which is then used to generate a feasible trajectory for the AM. 
% (4) {Placement Execution:} The AM follows the planned trajectory to place the object at the predicted location. 
% Bottom: Hardware demonstrations of language-guided aerial placement tasks across diverse scenarios. *An electric cable is connected to drone for power supply.}
\captionof{figure}{\small \textbf{Overview of AERMANI-PLACE.} Our method infers placement locations from language instructions and RGB-D observations using an Aerial Manipulator (AM). (1) \textit{Observation:} The AM receives the scene view and instruction. (2) \textit{Prediction:} An image editing model generates a 2D visual placement marker. (3) \textit{Recovery \& Formulation:} The marker is depth-projected to recover a 3D metric point and plan a feasible trajectory. (4) \textit{Execution:} The AM completes the placement. Bottom: Real-world hardware demonstrations (*drone tethered for power only ).}
    \label{fig:arch}
    
    \vspace{-4mm} % Space between caption and the start of the columns
\end{strip}

\begin{abstract}
Object placement is a fundamental component of aerial manipulation tasks, yet existing systems typically require the desired placement position to be specified explicitly in metric coordinates. Such interfaces are not intuitive and require users to reason about coordinate frames and scene geometry, making them difficult to use in practical deployments. In contrast, humans often communicate spatial goals through a combination of language and pointing gestures. Inspired by this observation, we present AERMANI-PLACE, a framework for language-guided object placement with aerial manipulators. Given a scene image and a natural language instruction, an image editing model generates a modified version of the scene containing a visual marker that indicates where the object should be placed. This marker is then grounded into the physical environment using depth observations to recover a metric place point, after which a placement trajectory is generated and executed by the aerial manipulator. We evaluate the proposed approach on a test set of 100 language-guided placement tasks and demonstrate successful execution on a real aerial manipulation platform. Experimental results show that the proposed method reliably infers placement locations from language instructions with an average success rate of 87\% on the test-set and transfers effectively to real-world aerial manipulation with an average success rate of 72\%.

{\color{blue}Video: https://youtu.be/SgwwgLBsv0g}
\end{abstract}

\section{Introduction}

Aerial manipulators (AMs) combine the mobility of aerial robots with the dexterity of robotic arms, enabling physical interaction with objects in locations that are difficult or impossible to reach with ground-based manipulators. This capability supports applications such as infrastructure servicing, inspection, warehouse operations, and disaster response \cite{yadav2024modular}. Although substantial progress has been made in AM control and task-specific primitives such as aerial grasping \cite{Ubellacker24npj-softDrone3, yadav2025integrated, Kim2016_visionGuidedAerial}, the reverse task of object placement remains comparatively underexplored, despite being essential for completing manipulation workflows.

Most existing AM systems assume that the desired placement position is explicitly specified in advance. This typically requires a user or operator to provide precise target coordinates, after which a trajectory is generated for execution. However, metric coordinate specification is unintuitive and error-prone, requiring users to reason about coordinate frames and scene geometry, where small errors may cause task failure or safety risks. In many real-world settings, obtaining such placement poses is also impractical.

Humans avoid explicit metric specification by combining language with deictic gestures, such as saying \textit{“put the tool there”} while pointing. Inspired by this, we consider interfaces where an operator or high-level planner specifies placement locations directly within the robot’s visual field. Click-based interfaces reduce coordinate-entry friction but still require continuous human involvement. Greater autonomy requires the system to infer placement locations from high-level instructions. Recent work in ground-based manipulation has explored language-to-spatial grounding using Vision-Language-Action models and keypoint reasoning frameworks \cite{yuan2025robopoint, zhao2025anyplace, liu2024moka}; however, these approaches have not yet transferred effectively to AMs, primarily due to high latency.

To address this gap, we introduce \textbf{AERMANI-PLACE}, a training-free framework for language-guided object placement with AMs. Given an RGB scene image and a natural language instruction, our method leverages the spatial reasoning capabilities of off-the-shelf image editing models to generate a small visual marker at the intended placement location. This marker serves as a semantic anchor, which is grounded into 3D using onboard depth information. By reformulating placement as a visual “pointing” task, the method removes the need for predefined placement coordinates.

This work focuses on predicting the 3D placement position, namely the point where the AM gripper should release the object. While full 6-DoF orientation is important for assembly tasks, reliable orientation prediction from a single viewpoint remains challenging due to aerial instability and depth noise. Establishing robust language-to-position grounding therefore provides a practical foundation for more autonomous aerial pick-and-place workflows.

We evaluate AERMANI-PLACE on a 100-task test set covering diverse objects and indoor configurations, and further demonstrate the complete system on a physical AM platform. The results show that the method can reliably interpret placement instructions and execute placements in real-world settings.
Our main contributions are:
\begin{itemize}
\item A training-free framework for language-guided aerial placement that uses image editing models to generate visual pointing cues.
\item A geometric grounding pipeline that converts 2D visual markers into actionable 3D metric configurations for motion planning.
\item Experimental validation on 100 indoor tasks and an end-to-end real-world demonstration on an AM platform.
\end{itemize}

\section{Related Work}

Prior work on aerial manipulators (AMs) has studied multi-rotor platforms equipped with lightweight arms and grippers \cite{yadav2024modular}, demonstrating aerial grasping, tool interaction, and object transportation \cite{Kim2016_visionGuidedAerial,yadav2025integrated,sharma2025impedance,Ubellacker24npj-softDrone3}. Much of this work focuses on control, stabilization, and grasping under coupled aerial-manipulator dynamics, where precise perception and stable flight are critical.

Autonomous aerial grasping has been achieved using onboard sensing and visual feedback \cite{Ubellacker24npj-softDrone3}. However, object placement remains less explored. Existing approaches typically assume predefined metric placement poses, with motion planning and control used to execute trajectories to known targets \cite{torrente2021dataDrivenMPC,schulman2014trajopt}.

To reduce reliance on exact analytical models, recent studies have learned residual dynamics~\cite{cao2024computation, ujjawal2026learn, yadav2026learning, ujjawal2025aermani, yadav2025arcade, yadav2026physics}. Recent works also use language and vision models for grasping, placement, and task-level reasoning, including clutter-aware aerial grasping, language-grounded placement, and VLM-based skill selection~\cite{singh2026aerograb,mishra2026aeroplace, mishra2025aermani}. VLMs and LLMs have been used to interpret instructions, reason about scenes, and guide robot behavior \cite{saycan,rt2,driess2023palme}, while related methods ground language in visual observations to identify task-relevant objects or regions \cite{peract,zeng2020transporter}. Large-scale visual policy learning has also advanced manipulation \cite{graspnet,contactgraspnet,mahler2017dexnet,diffusionpolicy}, but often requires extensive task-specific data, which is difficult to obtain for specialized AM hardware.

A closely related direction grounds language into spatial affordances or keypoints for manipulation targets \cite{kpam,yuan2025robopoint,ye2026st4vla,chen2025bridgevla}. \textbf{RoboPoint} \cite{yuan2025robopoint} and \textbf{AnyPlace} \cite{zhao2025anyplace} map high-level instructions to image keypoints or 3D placement configurations, while \textbf{Moka} \cite{liu2024moka} uses mark-based visual prompting for open-vocabulary affordance grounding. Although effective for ground manipulators, these methods often depend on heavy VLA backbones or specialized training, limiting their suitability for AMs with unique viewpoints, latency constraints, and aerodynamic coupling.

Generative models have also been used to produce subgoal images or actionable flows for manipulation policies \cite{ni2024generate,image_visual_planner,novaflow,dharmarajan2025dream2flow}. In contrast, \textit{AERMANI-PLACE} uses image editing models to directly infer placement cues in the AM camera view. By predicting a visual marker at the desired location, it offers a training-free, interpretable alternative to VLA-based policies for grounding language into actionable aerial placement points.

\section{Methodology}
\begin{figure*}[h]
\centering
\includegraphics[width=1.0\textwidth]{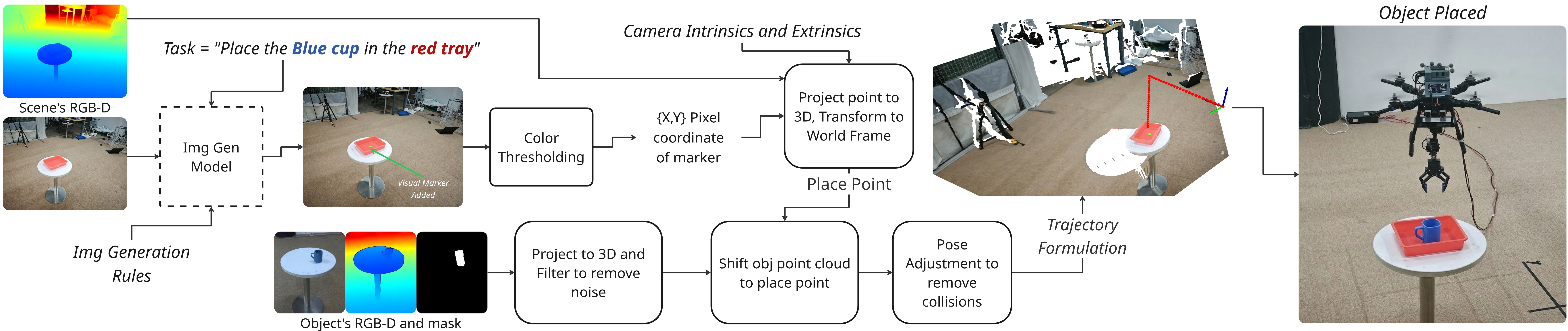}
% \caption{\small \textbf{Overview of the AERMANI-PLACE pipeline.}
% Given a natural language instruction and RGB-D observations of the placement scene, the system predicts a placement location and executes the task using an AM. 
% The scene image and instruction are provided to an image editing model, which generates a modified scene containing a small visual placement marker indicating where the object should be placed. 
% The marker is detected through color thresholding to recover its pixel coordinates, which are then projected into the scene using the RGB-D observations and camera calibration to obtain a metric placement point in the world frame. 
% The object geometry is reconstructed from its RGB-D observation and aligned with the predicted placement point. 
% A local pose adjustment step refines the placement configuration to remove collisions with the scene geometry. 
% Finally, a trajectory is generated for the AM to transport the object to the predicted location and complete the placement task.}
\caption{\small \textbf{Overview of the AERMANI-PLACE pipeline.} Given a language instruction and RGB-D scene observations, an image editing model generates a visual placement marker. The marker's pixel coordinates are extracted via color thresholding and depth-projected to recover a 3D metric point in the world frame. The object's geometry is reconstructed from its RGB-D view, aligned to the predicted point, and refined via local pose adjustment to prevent scene collisions. Finally, a trajectory is generated for the AM to execute the placement.}

\label{fig:imgflow}
\vspace{-4mm}
\end{figure*}

% Our approach enables aerial manipulators to execute object placement tasks specified through natural language in static environments. 
% The key idea is to first infer the intended placement location in image using a language-conditioned image editing model, and then recover the corresponding position in metric 3D space using depth observations. 
% The inferred placement location is subsequently adjusted to ensure collision-free placement before being executed by the aerial manipulator.

% Sec.~III-A introduces the problem formulation and provides an overview of the proposed pipeline. 
% Sec.~III-B describes how the placement location is predicted in image space using a language-guided visual marker. 
% Sec.~III-C explains how this marker is grounded into the 3D scene to determine a placement point for the object. 
% Sec.~III-D presents a local adjustment procedure to avoid collisions with the environment. 
% Finally, Sec.~III-E describes how the AM executes the placement using standard motion planning and control.

\subsection{Problem Formulation and Overview}

We consider the task of placing a grasped object in a static environment using an aerial manipulator (AM) based on a natural language instruction $\mathcal{L}$. The system receives an RGB-D observation of the placement scene $S = \{I, D\}$, where $I \in \mathbb{R}^{H \times W \times 3}$ is the color image and $D \in \mathbb{R}^{H \times W}$ is the registered depth map. We assume the camera intrinsic matrix $K$ and the camera pose $T_{cam} \in SE(3)$ in the world frame $W$ are known. The object is assumed to be stably held in the gripper using established grasping primitives.

The objective is to find a physically feasible placement point $P^* = (x, y, z) \in \mathbb{R}^3$ that satisfies the semantic constraints of $\mathcal{L}$. We decompose the framework into four stages:

\begin{enumerate}
    \item \textbf{Language-Guided Goal Localization:} A generative image-editing model $\Phi$ maps the input $\{I, \mathcal{L}\}$ to an edited image $I' = \Phi(I, \mathcal{L})$. $I'$ contains a visual marker (e.g., a green dot) at pixel coordinates $u \in \mathbb{R}^2$, representing the 2D projection of the desired placement location.
    \item \textbf{3D Placement Recovery:} The pixel $u$ is projected into the 3D world frame using the depth $D(u)$ and camera parameters to obtain the metric placement point $P_{init} = T_{cam} \cdot \pi(u, D(u), K)$, where $\pi$ denotes the projection function.
    \item \textbf{Collision-Aware Adjustment and Execution:} To ensure physical feasibility, a local refinement function $f: P_{init} \to P^*$ first resolves potential collisions with the environment geometry extracted from $D$. A motion planner then generates a trajectory $\mathcal{T}$ to move the AM from its current state $X_{curr}$ to the release configuration $X_{place}$ corresponding to the refined point $P^*$.
    % \item \textbf{Collision-Aware Refinement:} To ensure physical feasibility, a local refinement function $f: P_{init} \to P^*$ is applied to resolve potential collisions between the object and the environment geometry extracted from $D$ due to errors in depth.
    % \item \textbf{Placement Execution:} Finally, a motion planner generates a trajectory $\mathcal{T}$ to move the AM from its current state $X_{curr}$ to the release configuration $X_{place}$ corresponding to the refined point $P^*$.
\end{enumerate}

\subsection{Language-Guided Goal Localization}

Mapping high-level semantic instructions $\mathcal{L}$ to metric coordinates is challenging. We bridge this using a 2D-centric approach, employing a generative image-editing model $\Phi$ as a zero-shot spatial reasoner. Given a scene image $I$ and instruction $\mathcal{L}$, $\Phi$ generates an edited image $I'$ containing a high-contrast visual marker (e.g., a neon green dot) at the target site. This leverages the model's internal spatial knowledge without requiring explicit 3D training data. 

To preserve the physical environment's fidelity, a structured prompt (Listing~\ref{lst:prompt}) enforces \textit{Consistency Constraints}. $\Phi$ is instructed to maintain the original viewpoint, lighting, and geometry to prevent disruptive hallucinations. Finally, we extract the marker's pixel coordinates $u = (u, v)$ via HSV color-thresholding, yielding a 2D goal hypothesis anchored directly to the scene's visual features.

\begin{lstlisting}[language=bash, caption={Structured prompt for goal localization.}, label={lst:prompt}, basicstyle=\ttfamily\footnotesize, breaklines=true, frame=single]
You are given a scene image. 
The goal is to indicate where an object should be placed according to the instruction.
Instruction: [USER_INSTRUCTION_HERE]
Edit the scene image by adding a small neon green dot on the surface where the object should be placed.
Requirements:
- Keep the scene geometry and objects unchanged.
- Do not modify lighting or textures.
- Do not modify camera pov.
- Add only the small green placement marker.
\end{lstlisting}

\subsection{3D Placement Recovery}

The 2D marker provides a semantic hypothesis of the goal in the image domain. To translate this into an actionable robotic command, we ground the pixel coordinates $u = (u, v)$ into the 3D world frame $W$ and align the object geometry accordingly.

\textbf{Point Grounding:} Given the registered depth map $D$, we retrieve the depth value $d = D(u)$ at the marker's centroid. Using the camera intrinsic matrix $K$, the 3D position in the camera frame $P_{cam}$ is recovered via back-projection. This is then transformed into the world frame using the known camera pose $T_{cam}$:
\begin{equation}
    P_{init} = T_{cam} \cdot \begin{bmatrix} (u - c_x) \frac{d}{f_x} \\ (v - c_y) \frac{d}{f_y} \\ d \\ 1 \end{bmatrix}
\end{equation}
where $(f_x, f_y)$ and $(c_x, c_y)$ are the focal lengths and principal point, respectively.

\textbf{Object Geometry Reconstruction:} To ensure the object is placed realistically on the surface, we utilize its pre-grasp RGB-D observation. We apply a binary segmentation mask $M_O$ to isolate the object and back-project the masked depth pixels to generate a source point cloud $\mathcal{C}_O$. We then identify the set of base support points $\mathcal{C}_{base} \subset \mathcal{C}_O$ by extracting points with the lowest vertical coordinates in the object’s local frame.

\textbf{Surface Alignment:} The placement goal $P_{init}$ is treated as the target point for the object's base. We compute a translation vector $t$ that shifts  $\mathcal{C}_{base}$ to $P_{init}$. By aligning the object's "lowest" geometry with the predicted scene point, we ensure that the AM's release configuration accounts for the physical extents of the payload. This produces an initial 3D placement configuration $X_{init}$, which serves as the input for the subsequent collision-avoidance and motion planning stages.

\subsection{Collision-Aware Adjustment and Execution}

To compensate for depth sensor noise and 2D marker inaccuracies, we perform a local $xyz$ geometric refinement of the initial placement point $P_{init}$. We first optimize the height ($z$-axis) by evaluating the object point cloud $\mathcal{C}_O$ at $P_{init}$. The object is shifted upward until collision-free, then lowered to the point of first contact. This "touch-down" approach ensures placement on the physical surface despite offsets in the depth map $D$. If lateral collisions persist, a local horizontal ($xy$-plane) search identifies $P^*$, the closest collision-free position to the original prediction. 

Once $P^*$ is established, the AM executes a multi-stage \textit{top-down} trajectory to deliver the payload. A global planner routes the AM from its current state $X_{curr}$ to an \textit{approach pose} at a fixed vertical offset $h_{offset}$ directly above $P^*$, avoiding lateral obstacles. The AM then performs a controlled vertical descent to minimize aerodynamic disturbances such as downwash. Upon reaching the release altitude:
\begin{equation}
    z_{release} = z_{P^*} + \delta
\end{equation}
where $\delta$ is a small safety margin, the gripper releases the object. Finally, a vertical retraction climb is executed to clear the placement site, mitigating ground-effect instability and ensuring the object remains aligned with the semantically intended location.

\begin{figure}[t]
\centering
\includegraphics[width=0.5\textwidth]{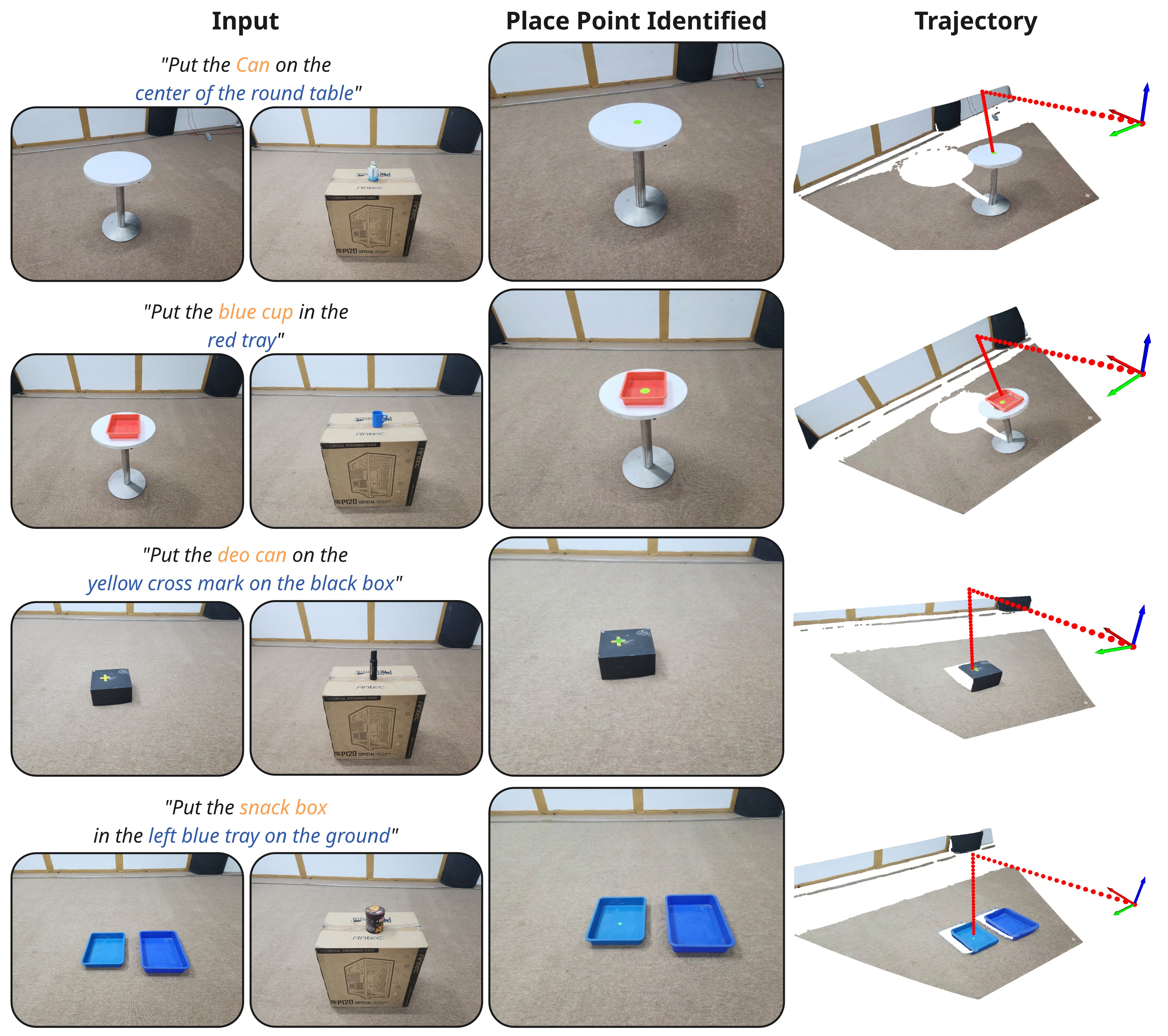}
\caption{\small \textbf{Qualitative placement results.} Rows illustrate distinct tasks across difficulty categories. Left to right: \textit{Input} shows observations and instruction $\mathcal{L}$; \textit{Place Point} displays the marker generated by $\Phi$; and \textit{Trajectory} visualizes the 3D approach path. The framework successfully handles complex relative positioning, precise feature targeting, and semantic disambiguation.}
% \caption{\small \textbf{Qualitative results from the language-guided placement benchmark.}
% Each row illustrates a distinct placement task across our difficulty categories. 
% From left to right: {Input} presents the scene and object observations alongside the natural language instruction $\mathcal{L}$; 
% {Place Point Identified} shows the semantic marker generated by the image editing model $\Phi$; 
% and {Trajectory} displays the 3D grounded scene with the planned approach and descent path for the AM. 
% The examples demonstrate the framework's ability to resolve complex instructions, including relative positioning, precise feature targeting (e.g., ``yellow cross mark''), and semantic disambiguation (e.g., ``left blue tray'').}

\label{fig:benchtest}
\vspace{-5mm}
\end{figure}

\section{Experimental Evaluation}

We evaluate the \textit{AERMANI-PLACE} framework through two distinct studies: a large-scale test-set of 100 language-guided placement tasks and real-world hardware demonstrations using an aerial manipulation platform.

\subsection{Test-set and Setup}

To systematically evaluate the robustness of our pipeline, we constructed a test set of 100 diverse placement tasks captured in a laboratory environment. Each task consists of:
\begin{enumerate}
    \item \textbf{Observations:} A pair of RGB-D images representing the object $O$ (pre-grasp) and the placement scene $S$.
    \item \textbf{Instruction $\mathcal{L}$:} A natural language command specifying the spatial goal (e.g., ``place the battery in the red bin'').
    \item \textbf{Ground Truth:} A manually annotated 3D coordinate $P_{GT}$ representing the optimal placement site.
\end{enumerate}

The tasks span four difficulty categories: \textit{Direct Surface} (on a table), \textit{Relative Positioning} (next to an object), \textit{Semantic Disambiguation} (left blue tray), and \textit{Variable Texture Surfaces}. Examples are shown in Fig \ref{fig:benchtest}.

\subsection{Hardware Configuration}

Our custom aerial manipulation platform consists of a quadrotor equipped with a lightweight robotic arm ($4$ Degrees of Freedom) and a gripper. 
\begin{itemize}
    \item \textbf{Perception:} Onboard RGB-D sensing is provided by a ZED2-i camera.
    \item \textbf{Computation:} State estimation and low-level control run on an onboard NVIDIA Jetson Nano Super.
    \item \textbf{Localization:} High-accuracy state estimation is provided by a Optitrack motion capture system.
    \item \textbf{Models used:} Nano banana pro model is used for experiments, object masks are acquired with SAM3\cite{carion2025sam3segmentconcepts}.
\end{itemize}

% \subsection{Baselines and Compared Models}

% We compare our framework against two human-centric baselines:
% \begin{itemize}
%     \item \textbf{Ground Truth (GT):} An oracle baseline where $P^*$ is set to $P_{GT}$. This defines the upper bound of the geometric pipeline.
%     \item \textbf{User-Click:} A human operator selects the target pixel $u$ in a live camera view, bypassing the image-editing model $\Phi$.
%     \item \textbf{VLA-Point Baseline (VLA-Base):} 
% Following recent work in language-to-point grounding (e.g., RoboPoint), we compare against a VLA-centric approach. While effective, such models are typically trained on stationary, ground-based datasets. In our experiments, we observe that these models struggle with the unique perspectives and sensor noise of aerial platforms, whereas \textit{AERMANI-PLACE} utilizes general-purpose generative priors that generalize better to unstructured aerial views.
% \end{itemize}

% \textbf{Note on Dismissed Baselines:} 
% We intentionally exclude end-to-end trained policies (e.g., BridgeVLA, ST4VLA) from our comparison. These methods require massive domain-specific datasets and lack the modular 3D geometric grounding necessary for safe aerial execution in diverse, unmapped environments.

% To evaluate the impact of the generative backbone, we test the proposed framework using five state-of-the-art image editing models: \textit{Google Nano Banana Pro}, \textit{Qwen-VL-Edit}, \textit{FLUX-Kontext}, \textit{GPT-4o-Image}, and \textit{OmniGen2}. All models utilize the identical prompt structure defined in Sec.~III-B.

\subsection{Baselines and Compared Models}

We compare \textit{AERMANI-PLACE} against two human-centric baselines and three state-of-the-art (SOTA) language-grounding frameworks and with 3 image editing models(Google nano banana pro, Qwen-VL-Edit and Kling):

\begin{itemize}
    \item \textbf{Ground Truth (GT):} An oracle baseline where $P^*$ is set to the manually annotated $P_{GT}$. This defines the theoretical upper bound for our geometric and control pipeline.
    \item \textbf{User-Click:} A human operator manually selects the target pixel $u$ in a live camera view. This serves as a benchmark for user-in-the-loop specify-and-place performance.
    \item \textbf{RoboPoint \cite{yuan2025robopoint}:} A VLA-based baseline that predicts image keypoints for spatial affordances. As a model trained on ground-based manipulation, it provides a comparison for cross-domain spatial reasoning.
    \item \textbf{Anyplace \cite{zhao2025anyplace}:} A general-purpose framework for object placement. We evaluate its ability to generalize to the unstructured viewpoints of an aerial platform.
    \item \textbf{Moka \cite{liu2024moka}:} A training-free baseline that utilizes mark-based visual prompting for open-vocabulary manipulation. This comparison highlights the difference between VQA-based marking and our generative image-editing approach.
\end{itemize}

The tests are conducted in a fair manner and ensure that all baselines receive the same information under the same conditions. The baselines have been implemented on local computer with high end GPU compute.

\subsection{Evaluation Metrics}

We evaluate the performance of \textit{AERMANI-PLACE} and the baselines using the following quantitative metrics:

\begin{itemize}
    \item \textbf{Marker Prediction Success (MPS):} The percentage of trials where the image editing model $\Phi$ correctly generates a detectable visual marker without altering the underlying scene geometry or lighting.
    \item \textbf{Placement Error (PE):} The Euclidean distance (in cm) between the predicted placement point $P^*$ and the manually annotated ground truth $P_{GT}$.
    \item \textbf{Placement Success (PS):} A binary metric indicating overall task completion. A trial is successful if the placement point is within 5~cm of the ground truth.
\end{itemize}

% \begin{table}[t]
% \centering
% \caption{Placement performance on the 100-task test set. The proposed method is evaluated using multiple image editing models.}
% \label{tab:results}
% \begin{tabular}{lcccc}
% \toprule
% \textbf{Method} & \textbf{MPS $\uparrow$} & \textbf{CFP $\uparrow$} & \textbf{PE(cm) $\downarrow$} & \textbf{PS $\uparrow$} \\
% \midrule
% Ground Truth (GT) & -- & 100 & 0.0 & 100 \\
% User Click Baseline & -- & 98 & 1.2 & 96 \\
% \midrule
% Ours (Nano Banana Pro) & 94 & 91 & 1.8 & 87 \\
% Ours (Qwen Image Edit) & 88 & 84 & 2.4 & 81 \\
% Ours (FLUX Kontext) & 82 & 78 & 2.9 & 74 \\
% Ours (GPT Image) & 75 & 71 & 3.4 & 68 \\
% Ours (OmniGen2) & 66 & 62 & 4.1 & 60 \\
% \bottomrule
% \end{tabular}
% \end{table}
\begin{table}[t]
\centering
\caption{Placement performance on the 100-task benchmark.}
\label{tab:results}
\scalebox{0.85}{
\begin{tabular}{lccccc}
\toprule
\textbf{Method} & \textbf{Training} & \textbf{Latency(s)} & \textbf{MPS $\uparrow$} & \textbf{PE(cm) $\downarrow$} & \textbf{PS $\uparrow$} \\
\midrule
Ground Truth (GT) & -- & -- & -- & 0.0 & 100 \\
User Click & -- & $<1.0$s & -- & 1.2 & 96 \\
\midrule
RoboPoint \cite{yuan2025robopoint} & Yes & $>1$min & 89 & 2.6 & 82 \\
Anyplace \cite{zhao2025anyplace} & Yes & $>1$min & 85 & 3.1 & 78 \\
Moka \cite{liu2024moka} & No & $\sim30.0$s & 87 & 2.8 & 80 \\
\midrule
\textbf{Ours (Nano Banana Pro)} & \textbf{No} & \textbf{$\sim12.0$s} & \textbf{94} & \textbf{1.8} & \textbf{87} \\
Ours (Qwen-VL-Edit) & No & $\sim25.0$s & 91 & 2.2 & 84 \\
Ours (Kling 1.5 Edit) & No & $\sim22.0$s & 89 & 2.4 & 82 \\
\bottomrule
\end{tabular}
}
\end{table}

\begin{figure}[t]
\centering
\includegraphics[width=0.35\textwidth]{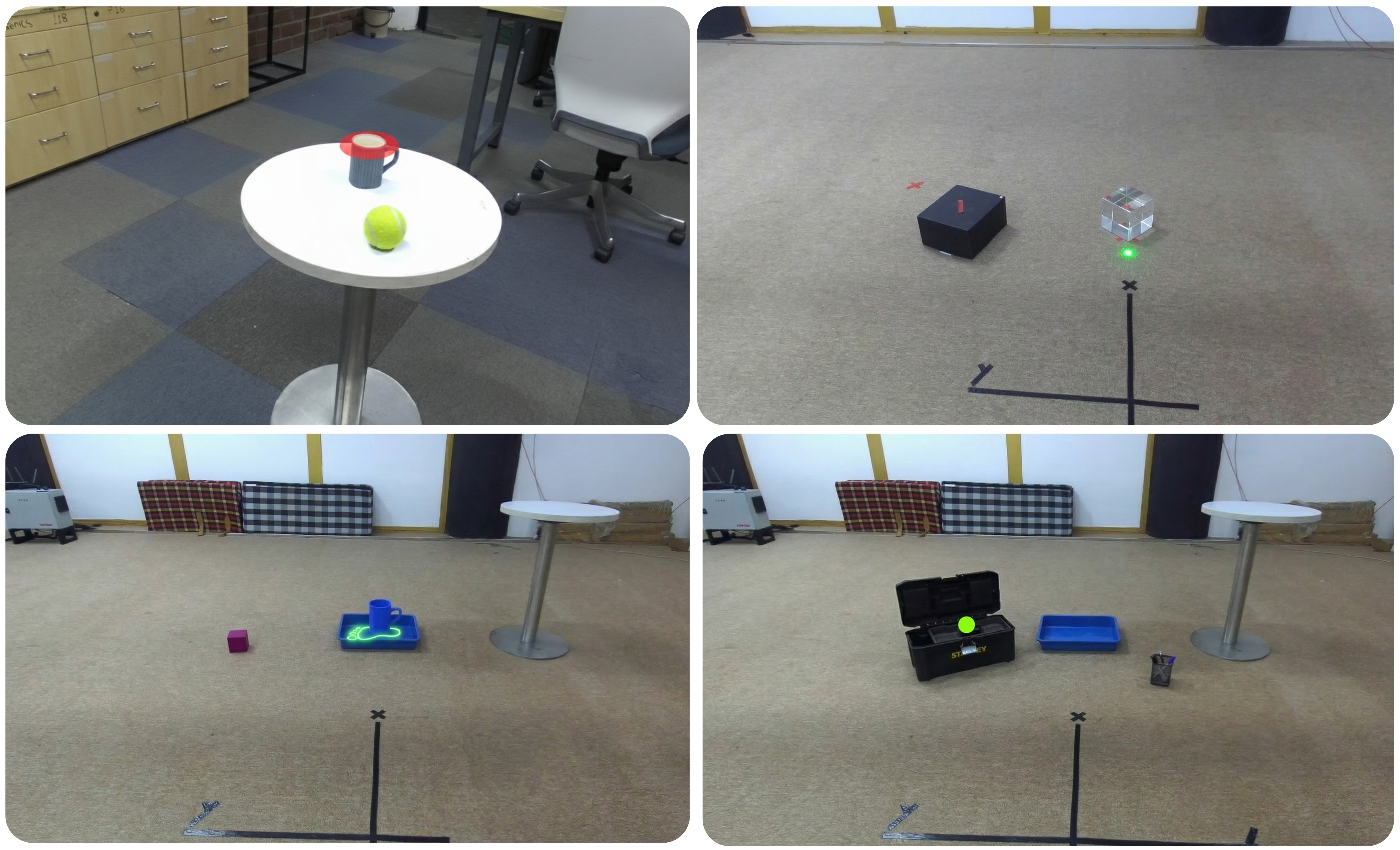}
% \caption{\small \textbf{Failure cases of placement marker generation.}
% Examples where the image editing model produces incorrect or ambiguous placement cues. 
% Top-left: The model introduces additional objects and predicts an incorrect marker location. 
% Top-right: The placement marker is not generated reliably, resulting in an ambiguous placement point. 
% Bottom-left: The model hallucinates the target object in the scene while generating the marker. 
% Bottom-right: The predicted marker is not aligned with the support surface, which would result in an unstable placement configuration.}
\caption{\small \textbf{Marker generation failure cases.} 
Incorrect or ambiguous cues from the image editing model. 
Top-left: Hallucinates extra objects and incorrect marker locations. 
Top-right: Unreliable or ambiguous marker generation. 
Bottom-left: Hallucinates the target object directly into the scene. 
Bottom-right: Marker is misaligned with the support surface, risking unstable placement.}

\label{fig:failure}
\vspace{-5mm}
\end{figure}

% \subsection{Results and Discussion}

% \subsubsection{Test-Set Performance and Baseline Analysis}
% As shown in Table~\ref{tab:results}, \textit{AERMANI-PLACE} achieves a competitive Placement Success (PS) of \textbf{87\%}, demonstrating performance that is comparable to established SOTA models such as \textbf{RoboPoint (82\%)} and \textbf{zhao2025anyplace (78\%)}. Significantly, our framework reaches these levels while remaining entirely \textbf{training-free}, offering a viable alternative to end-to-end policies that require extensive domain-specific datasets. The variations in baseline performance likely stem from the shift in perspective; while existing models are often optimized for ground-based, stationary viewpoints, our modular generative approach adapts naturally to the unstructured camera angles of an aerial platform. Furthermore, the similarity in metric precision—with our framework achieving a Placement Error of 1.8~cm compared to 2.6~cm in RoboPoint—suggests that general-purpose generative priors can provide a robust and efficient proxy for the specialized spatial reasoning found in trained VLA models.

\subsection{Results and Discussion}

\subsubsection{Test-Set Performance and Baseline Analysis}
As shown in Table~\ref{tab:results}, \textit{AERMANI-PLACE} achieves an \textbf{87\%} Placement Success (PS), comparable to trained SOTA models like \textbf{RoboPoint (82\%)} and \textbf{Anyplace (78\%)}. Crucially, our framework is entirely \textbf{training-free}. Baseline performance degradation likely stems from their reliance on stationary, ground-based viewpoints, whereas our generative approach seamlessly adapts to unstructured aerial perspectives. Additionally, our high metric precision (1.8~cm Placement Error vs. 2.6~cm for RoboPoint) demonstrates that general-purpose generative priors provide a robust, efficient alternative to the specialized spatial reasoning of VLA models without requiring domain-specific datasets.

\subsubsection{Failure Case Analysis}

We identify four primary failure modes in the \textit{AERMANI-PLACE} pipeline (Fig.~\ref{fig:failure}):

\begin{itemize}
    \item \textbf{Object Hallucination:} The generative model $\Phi$ inserts the target object into the scene alongside the marker (bottom-left), causing conflicting depth observations during grounding.
    \item \textbf{Scene Distortion:} The model violates consistency constraints by adding phantom objects or shifting surfaces (top-left). This causes high Placement Error (PE) as the grounded scene diverges from the physical environment.
    \item \textbf{Marker Misalignment:} The predicted marker floats "in mid-air" off a valid support surface (bottom-right). While $z$-axis refinement (Sec.~III-D) resolves small offsets, large misalignments cause unstable placements.
    \item \textbf{Ambiguous Prompt Interpretation:} Faint or indistinct markers (top-right) fail color-based detection thresholds, resulting in a Marker Prediction Failure (MPF).
\end{itemize}

These modes highlight that while training-free generative models are potent spatial reasoners, their probabilistic nature necessitates robust downstream geometric checks to ensure safe autonomous execution.

\begin{figure}[h]
\centering
\includegraphics[width=0.35\textwidth]{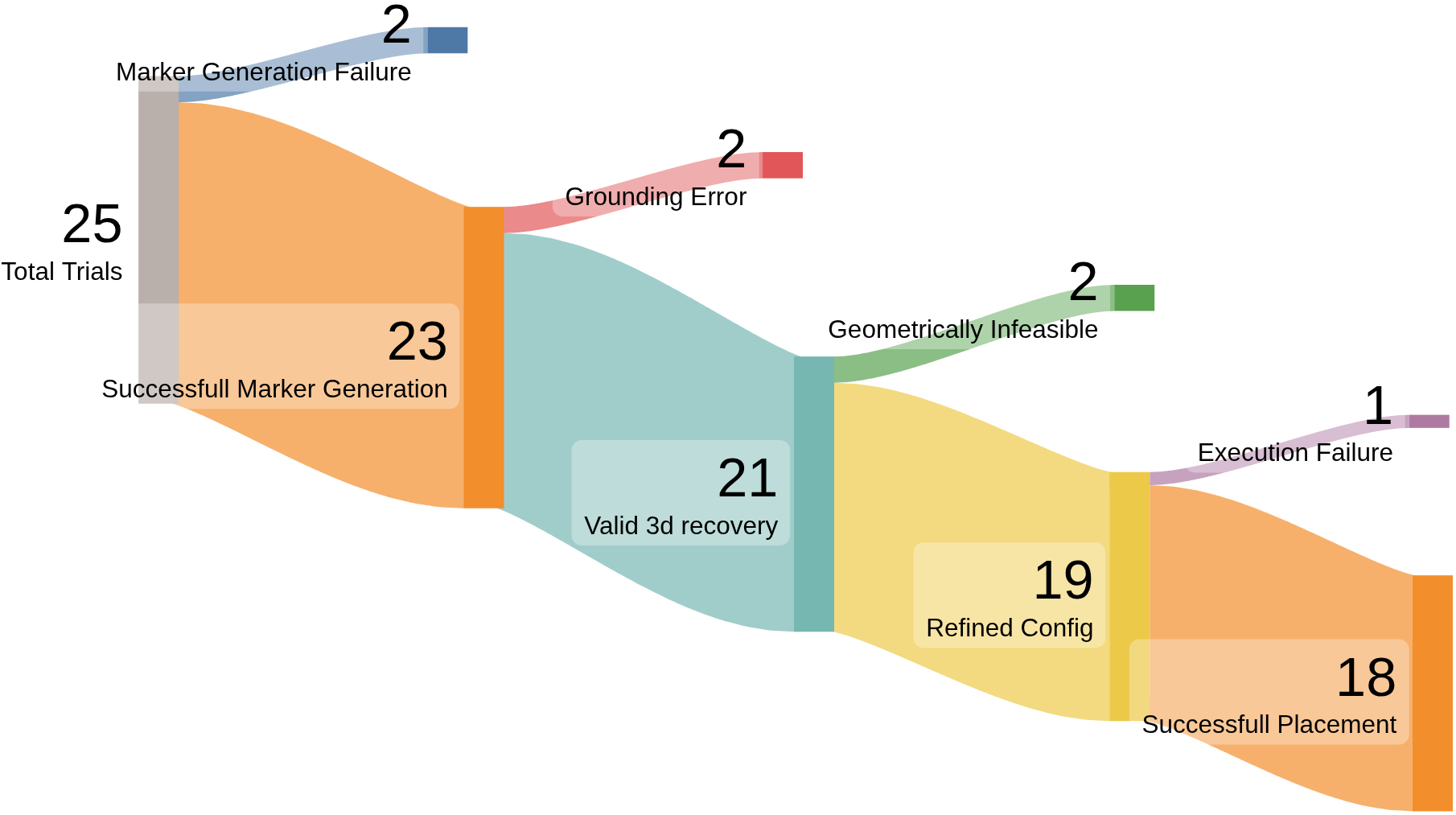}
\caption{\small \textbf{Failure analysis of hardware trials.} Stage-wise performance across 25 real-world trials. \textit{Semantic Localization} generated valid visual markers in 92\% of cases. \textit{Collision-Aware Adjustment} (Sec.~III-D) robustly recovered initially geometrically infeasible configurations. The 72\% final success rate accounts for losses from depth sensor errors, unresolved geometric constraints, and minor execution drift during aerial descent.}
% \caption{\small \textbf{Failure analysis of hardware trials.} 
% The diagram illustrates the stage-wise performance of AERMANI-PLACE across 25 real-world hardware trials. 
% The pipeline begins with \textit{Semantic Localization}, where 92\% of trials successfully generated a detectable visual marker. 
% Subsequent stages highlight the robustness of the \textit{Collision-Aware Adjustment} (Sec.~III-D) in recovering 3 cases that were initially geometrically infeasible. 
% The final success rate of 72\% accounts for losses due to depth sensor grounding errors, unresolved geometric constraints, and minor execution drift during the aerial descent phase}
\label{fig:hdw}
\vspace{-5mm}
\end{figure}

\subsubsection{Hardware Demonstration}

We validated the \textit{AERMANI-PLACE} framework on a physical aerial platform across 25 real-world trials spanning four difficulty categories, utilizing the best-performing Nano Banana Pro model. The system achieved a \textbf{72\%} success rate (18/25 successful placements). A stage-wise failure analysis (Fig.~\ref{fig:hdw}) reveals:

\begin{itemize}
    \item \textbf{Semantic Localization:} The image-editing model $\Phi$ failed to generate valid markers in 2 trials (8\%), primarily due to object hallucination.
    \item \textbf{Geometric Grounding:} Depth sensor noise (ZED2-i) caused 2 failures where $P_{init}$ projected into invalid space.
    \item \textbf{Refinement and Execution:} \textit{Collision-Aware Adjustment} (Sec.~III-D) resolved initial collisions in 21 cases. However, 2 trials failed when no local collision-free point $P^*$ existed, and 1 failed during descent due to aerodynamic oscillations causing object drift before release.
\end{itemize}

The \textit{top-down approach} (Sec.~III-E) proved vital for hardware stability, effectively preventing AM downwash from displacing lightweight objects. While $z$-axis optimization robustly handles minor depth offsets, improving generative consistency remains necessary to resolve localization bottlenecks in complex real-world scenes.

% \section{Conclusion}
% In this work, we presented \textbf{AERMANI-PLACE}, a training-free framework for language-specified object placement with aerial manipulators. By reformulating the placement task as a visual "pointing" problem, we eliminate the need for unintuitive metric coordinates and specialized training datasets. Our approach leverages the emergent spatial reasoning of off-the-shelf image editing models to generate semantic markers that are grounded into the physical environment through a robust geometric pipeline.Extensive evaluation on a 100-task benchmark and a physical aerial platform demonstrates the efficacy of our method, achieving success rates of 87\% and 72\%, respectively. A key finding of our hardware study is that while generative models provide a powerful interface for human intent, the integration of collision-aware geometric refinement and a "top-down" approach strategy is essential for mitigating sensor noise and aerodynamic disturbances inherent in aerial systems.Future research will focus on extending this framework to full 6-DoF orientation reasoning to enable precise assembly tasks. Additionally, we aim to explore the use of multimodal models to enable dynamic replanning and real-time obstacle avoidance in unstructured environments. We believe that bridging the gap between high-level vision-language reasoning and low-level aerial control is a critical step toward achieving intuitive and truly autonomous human-robot interaction in real-world scenarios.

\section{Conclusion}

In this work, we presented \textbf{AERMANI-PLACE}, a training-free framework for language-guided object placement with AMs. By reformulating the placement task as a visual ``pointing'' problem, we eliminate the need for unintuitive metric coordinates and specialized training datasets. Our approach leverages the emergent spatial reasoning of off-the-shelf image editing models to generate semantic markers that are grounded into the physical environment through a robust geometric pipeline. 

Extensive evaluation on a 100-task benchmark and a physical aerial platform demonstrates that our framework achieves performance comparable to state-of-the-art trained models, reaching success rates of 87\% and 72\%, respectively. A key finding of our hardware study is that while generative models provide a versatile training-free interface for human intent, the integration of collision-aware geometric refinement and a ``top-down'' approach strategy is essential for mitigating sensor noise and aerodynamic disturbances inherent in aerial systems. Future research will focus on extending this framework to full 6-DoF orientation reasoning to enable precise assembly tasks.

\bibliographystyle{IEEEtran}
\bibliography{root}

\end{document}